\documentclass[11pt, a4paper,logo,  copyright, nonumbering]{paddleocr}
\usepackage[numbers]{natbib}
\usepackage{dblfloatfix}
\usepackage[normalem]{ulem}
\usepackage{float}
\usepackage{tocloft}
\usepackage{caption}
\usepackage{dramatist}
\usepackage{xspace}
\usepackage{array}
\usepackage{pifont} %
\usepackage{multirow}
\usepackage{geometry}
\usepackage{makecell}
\usepackage{tcolorbox}
\usepackage{xltabular}
\usepackage{titlesec} 
\usepackage{longtable}
\usepackage[hang, flushmargin]{footmisc}
\usepackage{booktabs} 
\usepackage{xcolor}   
\usepackage[table]{xcolor}
\usepackage{etoolbox}
\usepackage{url}
\usepackage{changepage}
\definecolor{lightpink}{RGB}{255, 182, 193}
\definecolor{iccvblue}{rgb}{0.21,0.49,0.74}

\hypersetup{
    colorlinks=true,
    linkcolor=blue,
    citecolor=blue,
    filecolor=blue,
    urlcolor=blue,
    pdfpagemode=UseNone,
    pdfstartview=FitH
}
\usepackage{threeparttable} 
\usepackage{soul}
\usepackage{amsfonts}
\usepackage{amsmath}
\usepackage{amssymb}
\usepackage{lineno}
\usepackage{multirow}
\usepackage{adjustbox}

\usepackage{CJKutf8}
\usepackage{subcaption}
\usepackage{setspace}

\usepackage{dsfont}
\usepackage{array} %
\usepackage{tabularx} %
\usepackage{xcolor} %
\usepackage{tabularx}
\usepackage{booktabs}

\usepackage{lipsum}  %
\usepackage{multicol} %
\usepackage{makecell} 
\usepackage{listings}
\usepackage{xcolor}
\usepackage{eccvabbrv}

\usepackage{tablefootnote}
\usepackage{hyperref}

\lstdefinelanguage{json}{
    basicstyle=\ttfamily\footnotesize,
    numbers=left,
    numberstyle=\tiny\color{gray},
    stepnumber=1,
    numbersep=8pt,
    showstringspaces=false,
    breaklines=true,
    frame=lines,
    backgroundcolor=\color{gray!10},
    morestring=[b]",
    literate=
     *{0}{{{\color{black}0}}}{1}
      {1}{{{\color{black}1}}}{1}
      {2}{{{\color{black}2}}}{1}
      {3}{{{\color{black}3}}}{1}
      {4}{{{\color{black}4}}}{1}
      {5}{{{\color{black}5}}}{1}
      {6}{{{\color{black}6}}}{1}
      {7}{{{\color{black}7}}}{1}
      {8}{{{\color{black}8}}}{1}
      {9}{{{\color{black}9}}}{1}
}

\makeatletter
\def\@BTrule[#1]{%
  \ifx\longtable\undefined
    \let\@BTswitch\@BTnormal
  \else\ifx\hline\LT@hline
    \nobreak
    \let\@BTswitch\@BLTrule
  \else
     \let\@BTswitch\@BTnormal
  \fi\fi
  \global\@thisrulewidth=#1\relax
  \ifnum\@thisruleclass=\tw@\vskip\@aboverulesep\else
  \ifnum\@lastruleclass=\z@\vskip\@aboverulesep\else
  \ifnum\@lastruleclass=\@ne\vskip\doublerulesep\fi\fi\fi
  \@BTswitch}
\makeatother

\addto\extrasenglish{
}

 {\begin{list}{}%
         {\setlength{\leftmargin}{#1}}%
         \item[]%
 }
 {\end{list}}

\reportnumber{001} %

\renewcommand{\today}{}

\title{\centering RT-DocLayout: Real-Time End-to-End Document Layout Analysis with Reading Order in the Wild}

\author[1,*]{
\small
Cheng Cui, Tingquan Gao, Xueqing Wang, Changda Zhou, 
\vspace{-0.3cm}
\\
\small
Hongen Liu, Ting Sun, Yubo Zhang, Zelun Zhang, Jiaxuan Liu,
\vspace{0.1cm}
\\
\small
Manhui Lin, Yue Zhang, Suyin Liang, Yiqing Xiang, Yi Liu 
\vspace{0.2cm}
\\
\small
\textbf{PaddlePaddle Team, Baidu Inc.} 
\\
\small
\texttt{paddleocr@baidu.com}
\vspace{-0.5cm}
}

\renewcommand{\phi}{\varphi}

\renewcommand{\leq}{\leqslant}

\renewcommand{\epsilon}{\varepsilon}
\renewcommand{\imath}{\mathrm{i}}

\newlength{\restsubwidth}
\newlength{\restsubheight}
\newlength{\restsubmoreheight}
\newcommand{\rest}[2]{%
        \settowidth{\restsubwidth}{\ensuremath{#2}}
        \settoheight{\restsubheight}{\ensuremath{{}_{#2}}}
        \ensuremath{{#1\hskip 0.5pt}_{\vrule\kern2pt\parbox[b][%
        4pt][b]{\the\restsubwidth}{%
                        \ensuremath{{}_{#2}}}}}
        }

\begin{abstract}
\vspace{-0.3cm} 
Accurate document layout analysis remains a critical bottleneck for document parsing systems, due to the intricate coupling among heterogeneous document layout elements, geometric distortions (\eg, paper warping and bending, perspective variations), and reading order within diverse layout structures. Existing approaches typically rely on fragmented multi-stage pipelines or computationally heavy generative Transformer architectures, leading to error propagation and limited efficiency.
    
    In this paper, we present RT-DocLayout, a highly efficient end-to-end framework for document layout analysis, designed as a front-end for document parsing tasks. The proposed model unifies classification, detection, pixel-level segmentation, and reading order prediction for layout elements within a single 33M-parameter architecture. Built upon the RT-DETR, our key contribution is a unified multi-task formulation within a single query-based decoder that simultaneously classifies, regresses bounding box, generates masks, and constructs relationship to reason reading order.
    
    By jointly learning geometric and structural representations, RT-DocLayout introduces multi-task optimization that substantially improves robustness under real-world document distortions. Extensive experiments on public benchmarks demonstrate \textbf{state-of-the-art performance in document layout analysis while maintaining real-time inference speed(132.1 FPS)}. When coupled with downstream OCR engines, RT-DocLayout significantly improves full-document reconstruction quality, providing a scalable and practical foundation for real-world document intelligence systems.
    

\end{abstract}

\begin{document}

\maketitle

\let\thefootnote\relax\footnote{%
    RT-DocLayout is released in PaddleOCR under the name PP-DocLayoutV3. RT-DocLayout is the name used in this academic publication, while PP-DocLayoutV3 is the name adopted in the PaddleOCR open-source project. The model weights are publicly available at \url{https://huggingface.co/PaddlePaddle/PP-DocLayoutV3}%
}
\vspace{0cm} 

\begin{figure*}[t]
  \centering

  \begin{subfigure}[t]{0.48\linewidth}
    \centering
    \includegraphics[width=\linewidth]{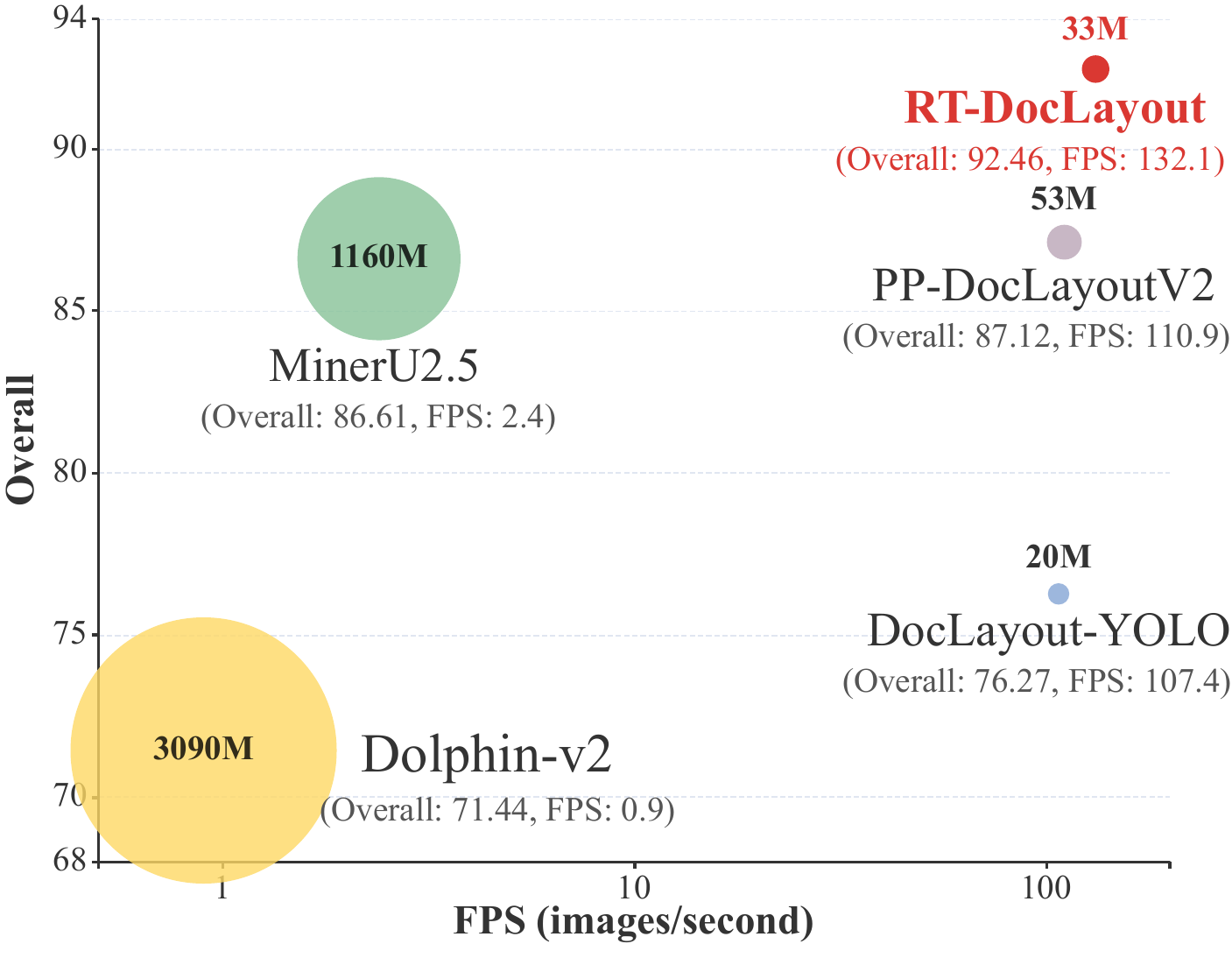}
    \caption{}
    \label{fig:teaser_a}
  \end{subfigure}
  \hfill
  \begin{subfigure}[t]{0.48\linewidth}
    \centering
    \includegraphics[width=\linewidth]{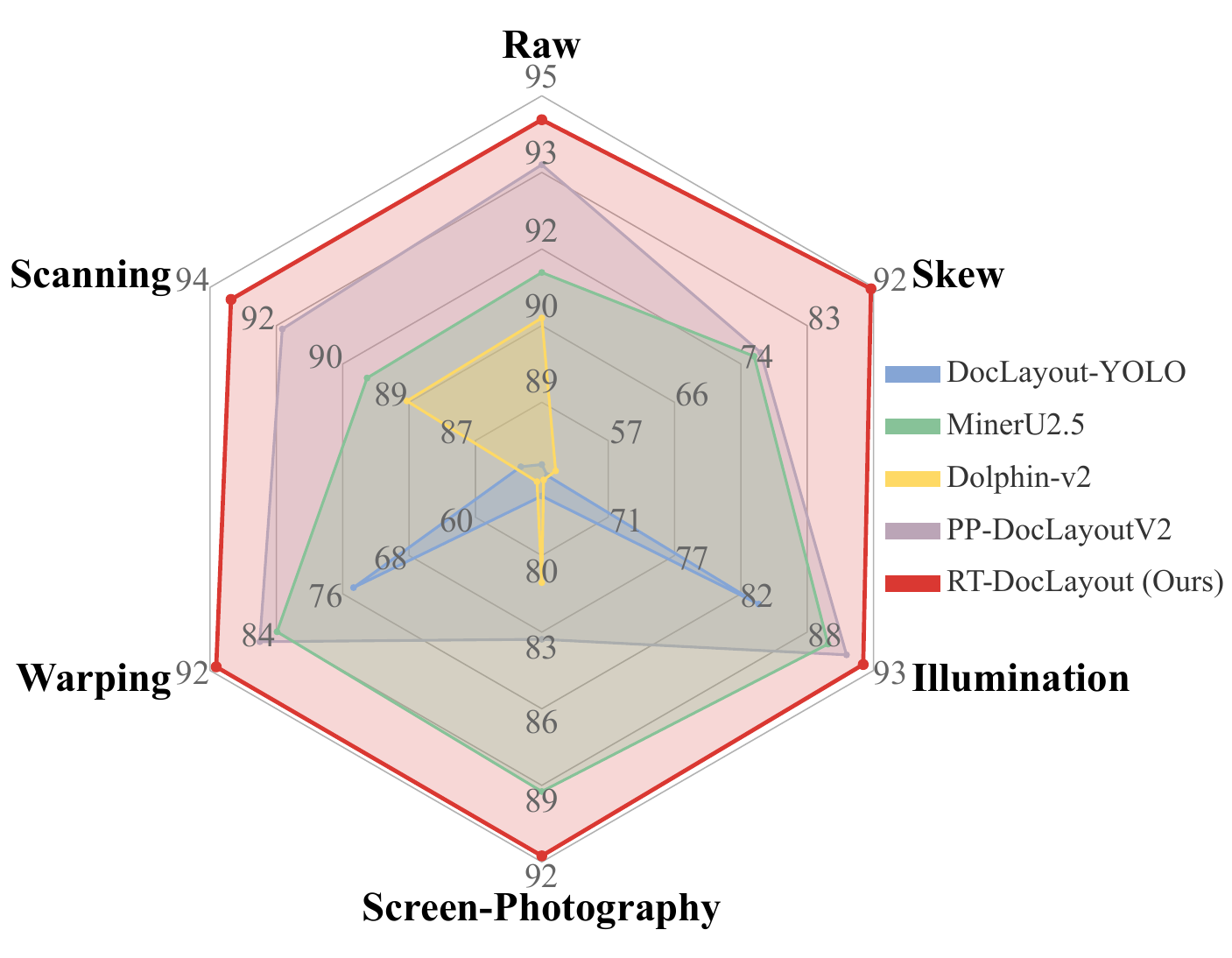}
    \caption{}
    \label{fig:teaser_b}
  \end{subfigure}
  
  \vspace{-3mm}
  
  \caption{
    \textbf{Performance and Robustness of RT-DocLayout.}
    Our proposed method achieves a new state-of-the-art in document layout analysis, demonstrating superior accuracy, real-time speed, and robustness. 
    \textbf{(a)} Comparison of performance (Overall Accuracy), inference speed (FPS), and model parameter count (indicated by bubble size). Our RT-DocLayout (red bubble) sets a new benchmark by achieving 92.46\% accuracy and 132.1 FPS with a compact 33M-parameter model, significantly surpassing previous leading methods like PP-DocLayoutV2 and MinerU2.5 in all three metrics. The overall accuracy is the average score across the six dimensions shown in (b). All experiments were performed using PaddleOCR-VL-1.5-0.9B as the downstream recognizer. Inference speed was tested on an NVIDIA A100 GPU with a batch size of 32. 
    \textbf{(b)} Head-to-head comparison on six challenging dimensions from the OmniDocBench v1.5 and Real5-OmniDocBench. RT-DocLayout consistently achieves the highest scores across all dimensions (Raw, Scanning, Warping, Screen-Photography, Illumination, and Skew), proving its exceptional robustness and generalizability under diverse real-world conditions.
  }
  
  \label{fig:teaser}
\end{figure*}
 
\section{Introduction}
\label{sec:intro}

Document Layout Analysis (DLA) is a fundamental prerequisite for automated Document Intelligence. Despite the success of deep learning in detecting layout elements, in complex scenarios involving mobile photography or digitization of bound archives, where geometric deformations are prevalent, document parsing is still hindered by two primary challenges: fragmented structural modeling and coarse geometric representation.

Current methodologies typically follow two suboptimal paths. First, cascaded pipelines attempt to reconstruct reading order by stacking a layout detector with secondary relationship models (\eg, GNNs~\cite{lee2021ropereadingorderequivariant} or heuristic sorting~\cite{ha1995recursive}). Such systems are inherently fragile, as detection errors propagate through the stages, and the lack of end-to-end integration prevents the model from capturing the synergistic dependencies between layout and topology. Second, large-scale pre-trained Transformers and VLMs (\eg, LayoutLMv3~\cite{huang2022layoutlmv3pretrainingdocumentai}, DiT~\cite{li2022ditselfsupervisedpretrainingdocument}, MinerU~\cite{niu2025mineru25decoupledvisionlanguagemodel}, Dolphin~\cite{feng2025dolphindocumentimageparsing}) have achieved impressive results but remain computationally expensive. Moreover, these models primarily rely on coarse-grained bounding boxes, which fail to provide the pixel-level masks necessary to isolate layout elements in warped, curved, or tilted documents. This lack of geometric precision often leads to content interference from background noise or adjacent elements, limiting the fidelity of downstream parsing.

To bridge this gap, we propose RT-DocLayout, a 33M-parameter unified framework based on the RT-DETR~\cite{zhao2024detrsbeatyolosrealtime} architecture. RT-DocLayout breaks the conventional boundary by unifying classification, detection, segmentation, and reading order prediction within a single query-based Transformer decoder. In a single forward pass, each object query concurrently classifies, regresses bounding boxes, generates high-fidelity masks for geometrically complex regions, and predicts reading order.

The core strength of RT-DocLayout lies in its ability to provide a multi-dimensional structural representation. 
By generating precise pixel-level masks instead of simple rectangles, the model provides a purified input for downstream OCR engines, effectively mitigating the interference caused by document deformations. 
This unified multi-task learning paradigm ensures that geometric shape and logical sequence are modeled within a shared feature space, leading to superior structural consistency. Notably, RT-DocLayout achieves state-of-the-art (SOTA) performance across multiple benchmarks with a compact 33M-parameter footprint, offering a highly efficient and scalable solution for real-time industrial document parsing, as shown on Fig.~\ref{fig:DataAug}.

Our contributions are summarized as follows:

\begin{itemize}
    \item We propose RT-DocLayout, a unified 33M-parameter Transformer framework for document layout analysis that achieves end-to-end classification, detection, pixel-level mask generation, and reading order prediction.
    
    \item We demonstrate that providing pixel-level masks for non-rigidly deformed regions significantly enhances the robustness of document parsing compared to traditional bounding-box-based methods.

    \item Extensive experiments demonstrate that RT-DocLayout achieves SOTA accuracy while maintaining real-time inference efficiency, validating its effectiveness as a practical front-end for document parsing systems.
\end{itemize}

\section{Related Work}
\label{sec:related}

\subsection{Document Layout Element Detection}

Early approaches adapted Faster R-CNN~\cite{ren2016fasterrcnnrealtimeobject} and Mask R-CNN~\cite{he2018maskrcnn} to document images~\cite{zhong2019publaynetlargestdatasetdocument, shen2021layoutparserunifiedtoolkitdeep}. DiT~\cite{li2022ditselfsupervisedpretrainingdocument} introduced self-supervised pre-training for document Transformers, and the LayoutLM series~\cite{xu2020layoutlm, xu2022layoutlmv2multimodalpretrainingvisuallyrich, huang2022layoutlmv3pretrainingdocumentai} pioneered joint text--layout--visual pre-training. DocLayout-YOLO~\cite{zhao2024doclayoutyoloenhancingdocumentlayout} improved inference speed for practical deployment. Despite these advances, all methods remain confined to axis-aligned bounding boxes that degrade OCR fidelity under geometric distortion, and none jointly models reading order.

\subsection{Reading Order Prediction}

Rule-based XYCut~\cite{ha1995recursive} is brittle on complex layouts. Data-driven methods cast the problem as sequence generation or relation prediction: LayoutReader~\cite{wang2021layoutreaderpretrainingtextlayout} pre-trains an autoregressive Transformer on the ReadingBank benchmark, and XYLayoutLM~\cite{gu2022xylayoutlmlayoutawaremultimodalnetworks} integrates spatial reading order into layout-aware pre-training. All these methods operate as post-hoc modules after a separate detector, forming cascaded pipelines prone to error propagation.

\subsection{Unified Document Parsing Frameworks}

Donut~\cite{kim2022ocrfreedocumentunderstandingtransformer} and Nougat~\cite{blecher2023nougatneuralopticalunderstanding} generate structured markup via OCR-free encoder--decoder models; UDOP~\cite{tang2023unifyingvisiontextlayout} further unifies vision, text, and layout in a generative framework. Large VLMs, including Dolphin~\cite{feng2025dolphindocumentimageparsing}, MinerU2.5~\cite{niu2025mineru25decoupledvisionlanguagemodel}, MinerU2~\cite{MinerU2}, and dots.ocr~\cite{li2025dotsocrmultilingualdocumentlayout}, extend this to multi-task document parsing at high computational cost. Closer to our setting, DLAFormer~\cite{wang2024dlaformerendtoendtransformerdocument} jointly predicts layout regions and reading order via a unified label space in an end-to-end DETR-based framework, and PP-DocLayoutV2~\cite{cui2025paddleocrvlboostingmultilingualdocument} appends a decoupled reading order module to RT-DETR~\cite{zhao2024detrsbeatyolosrealtime} for compact real-time parsing. Both remain restricted to axis-aligned bounding boxes and do not address geometric distortions from warping, perspective, or skew.

\textbf{Our approach.}~RT-DocLayout addresses all of the above limitations with a compact, purely visual architecture. Building on the insight that detection and segmentation can be jointly optimized via unified queries~\cite{li2022maskdinounifiedtransformerbased}, we extend RT-DETR~\cite{zhao2024detrsbeatyolosrealtime} to support end-to-end segmentation. Our unified multi-task formulation concurrently predicts bounding boxes, high-fidelity pixel-level masks, and reading order in a single forward pass, replacing both autoregressive generation and cascaded pipelines with deterministic, non-autoregressive decoding.

\section{Proposed Approach}
\label{sec:proposed_approach}

\subsection{Framework Overview}
\label{subsec:framework_overview}

Built upon the high-efficiency RT-DETR detector, RT-DocLayout reformulates document layout analysis as a unified, mask-centric prediction task. Instead of relying on conventional bounding-box representations, the framework produces pixel-accurate layout elements position while jointly modeling reading order relationships within a single Transformer architecture. This design enables reliable isolation of document layout elements under challenging conditions such as skewed or geometrically distorted pages, where bounding-box-based methods often introduce ambiguity and background interference. By integrating detection and reading order prediction into one framework, RT-DocLayout eliminates redundant post-processing stages and promotes structurally consistent and robust document layout analysis under geometric variations.

\subsection{Architecture Design}
\label{subsec:architecture_design}

\begin{figure}[ht]
\centering
\includegraphics[width=\linewidth]{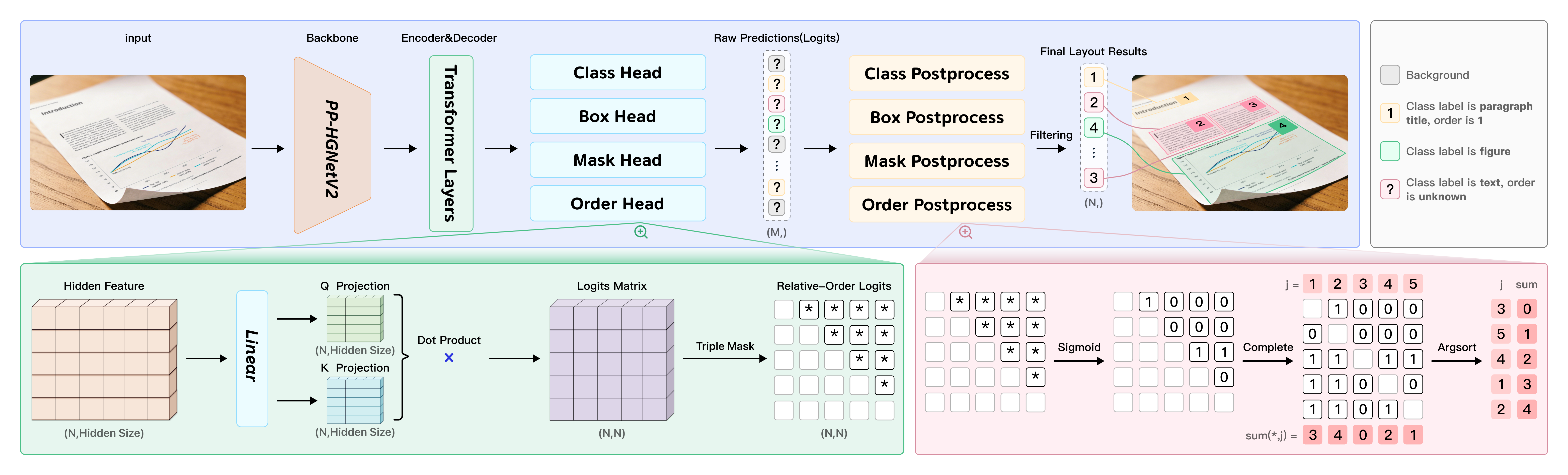}
\caption{The unified architecture of RT-DocLayout, featuring parallel heads for segmentation and reading order prediction.}
\label{fig:arch}
\end{figure}

Building upon the high-efficiency RT-DETR object detector~\cite{zhao2024detrsbeatyolosrealtime}, 
RT-DocLayout extends the detection paradigm toward unified structural prediction for document layout analysis, as shown on Fig.~\ref{fig:arch}.
Instead of relying solely on bounding-box representations, the model adds a mask-based detection head that produces pixel-accurate segments, enabling precise isolation of layout elements under challenging conditions such as skewed, curved, or geometrically distorted pages where axis-aligned boxes often introduce ambiguity and background contamination.

Unlike the decoupled pointer-network design used in prior systems~\cite{cui2025paddleocrvlboostingmultilingualdocument}, 
RT-DocLayout integrates reading order prediction directly into the Transformer decoder, allowing geometric perception and relationship representation of reading order to be optimized jointly. 
Following the query-based paradigm, the decoder iteratively refines $N$ object queries 
$Q=\{q_i\}_{i=1}^{N}\in\mathbb{R}^{N\times d}$, where each query encodes both spatial and structural semantics. 
Relative reading order logits matrix is calculated from the refined query embeddings, enabling absolute reading order for all elements by voting-based ranking strategy.

Specifically, refined queries are projected into a shared relational space to compute the pairwise precedence score:
\begin{equation}
    S_{i,j}=\frac{q_i^\top W_q^\top W_k q_j - q_j^\top W_q^\top W_k q_i}{\sqrt{d_h}},
\end{equation}
where $W_q, W_k \in \mathbb{R}^{d_h \times d}$ are learnable projection matrices that map queries to a relational embedding space, and $d_h$ denotes the head dimension. The resulting relation matrix $S\in\mathbb{R}^{N\times N}$ is inherently anti-symmetric ($S_{i,j}=-S_{j,i}$), where $S_{i,j}>0$ indicates that element $i$ precedes element $j$.

During inference, a voting-based ranking strategy converts pairwise relations into a globally consistent sequence. After applying the sigmoid function, the precedence vote for element $j$ is computed by summing over all incoming edges:
\begin{equation}
    V_j=\sum_{i=1}^{N}\sigma(S_{i,j}), \quad \text{where } S_{i,i}=0,
\end{equation}
where $V_j$ represents the expected number of elements preceding element $j$. The final reading order is obtained by sorting elements in ascending order of $V_j$—elements with fewer predecessors appear earlier in the sequence.

Through this joint optimization of classification, detection, segmentation, and pairwise order relationship within a single vision-centric architecture, 
RT-DocLayout produces the complete document structure in one forward pass, simultaneously outputting category labels, bounding boxes, pixel-level masks, and reading orders. 
This unified formulation promotes structurally consistent and geometrically robust document layout analysis while eliminating redundant post-processing and separated feature extraction stages.

\subsection{Loss Function Design}
\label{subsec:loss_function_design}

The training objective of RT-DocLayout is formulated as a multi-task optimization problem that jointly supervises classification, bounding-box, segmentation mask, and pairwise order relationship. We employ the Hungarian algorithm to establish optimal bipartite matching between predictions and ground-truth annotations before loss computation.

\textbf{Overall Objective.}
The total loss is a weighted combination of six components:
\begin{equation}
    \mathcal{L}_{\text{total}} = \sum_{k \in \mathcal{K}} \lambda_k \cdot \mathcal{L}_k,
\end{equation}
where $\mathcal{K} = \{\text{cls}, \text{bbox}, \text{giou}, \text{mask}, \text{dice}, \text{order}\}$ and the weights are set as $\lambda_{\text{cls}}=4$, $\lambda_{\text{bbox}}=5$, $\lambda_{\text{giou}}=2$, $\lambda_{\text{mask}}=5$, $\lambda_{\text{dice}}=5$, and $\lambda_{\text{order}}=50$. The reading order loss receives a substantially higher weight for two reasons. First, it addresses the gradient dilution problem: the order loss operates on $O(N^2)$ pairs, meaning each pairwise prediction receives gradients diluted by a factor of approximately $N/2$. In contrast, other object-level losses scale at $O(N)$, maintaining a more direct gradient impact. Second, while others losses benefit from deep supervision across all decoder layers (providing $L$ times more gradient updates for a model with $L$ layers), the order loss is computed only at the final layer and thus receives significantly fewer optimization signals throughout training.

\textbf{Classification and Detection Losses.}
For category prediction, we adopt focal loss with $\alpha=0.25$ and $\gamma=2.0$ to address class imbalance:
\begin{equation}
    \mathcal{L}_{\text{cls}} = -\frac{1}{N} \sum_{i} \alpha_{y_i} (1 - p_i)^{\gamma} \log(p_i),
\end{equation}
where $p_i$ denotes the predicted probability for the ground-truth class $y_i$. Bounding box regression combines $\ell_1$ loss and GIoU loss:
\begin{equation}
    \mathcal{L}_{\text{bbox}} = \|b_i - \hat{b}_i\|_1, \quad
    \mathcal{L}_{\text{giou}} = 1 - \text{GIoU}(b_i, \hat{b}_i),
\end{equation}
where $b_i$ and $\hat{b}_i$ represent predicted and ground-truth boxes respectively.

\textbf{Mask Segmentation Loss.}
Pixel-level mask supervision comprises binary cross-entropy and Dice loss:
\begin{equation}
    \mathcal{L}_{\text{mask}} = \text{BCE}(m_i, \hat{m}_i), \quad
    \mathcal{L}_{\text{dice}} = 1 - \frac{2|m_i \cap \hat{m}_i| + 1}{|m_i| + |\hat{m}_i| + 1}.
\end{equation}
Following MaskDINO, we employ point-based sampling with importance weighting to compute mask loss efficiently, sampling $K=12544$ points per mask with 75\% allocated to uncertain regions.

\textbf{Relative Reading Order Loss.}
The key contribution of our loss design is the pairwise reading order supervision. Given $N$ matched layout elements with ground-truth reading order indices $\{o_1, o_2, \ldots, o_N\}$, we construct a binary target matrix $T \in \{0, 1\}^{N \times N}$ where:
\begin{equation}
    T_{i,j} = \begin{cases}
        1 & \text{if } o_i < o_j \text{ (element $i$ precedes $j$)}\\
        0 & \text{otherwise}
    \end{cases}.
\end{equation}

The antisymmetric pairwise scorer predicts pairwise logits $S \in \mathbb{R}^{N \times N}$ with the constraint:
\begin{equation}
    S_{i,j} = \frac{f(q_i, q_j) - f(q_j, q_i)}{\sqrt{d_h}}, \quad f(q_i, q_j) = (W_q q_i)^\top (W_k q_j),
\end{equation}
ensuring $S_{i,j} = -S_{j,i}$ for consistent ordering relationships.

To emphasize local ordering correctness, we introduce locality-aware weighting. For element pairs within a dynamic neighbor window $k = \max(5, 0.3N)$, we apply a higher weight $w_{\text{local}}=2.0$:
\begin{equation}
    w_{i,j} = \begin{cases}
        w_{\text{local}} & \text{if } 0 < |o_i - o_j| \leq k\\
        1.0 & \text{otherwise}
    \end{cases}.
\end{equation}

To handle potential annotation noise, we employ Generalized Cross Entropy (GCE) loss with label smoothing ($\epsilon=0.01$):
\begin{equation}
    \mathcal{L}_{\text{order}} = \frac{1}{|P|} \sum_{(i,j) \in P} w_{i,j} \cdot \text{GCE}(S_{i,j}, T_{i,j}),
\end{equation}
where $P$ denotes the set of valid upper-triangular pairs (excluding diagonal), and $\text{GCE}(z, t) = (1 - p_t^q)/q$ with $p_t = \sigma(z)$ if $t=1$ else $1-\sigma(z)$, and $q=0.7$ for robustness.

\textbf{Deep Supervision.}
Auxiliary losses from intermediate decoder layers provide deep supervision for classification, detection, and segmentation, improving gradient flow and convergence. The reading order loss, however, is computed only at the final decoder layer. This design choice is motivated by the observation that accurate reading order prediction fundamentally depends on precise object localization and classification—intermediate layer representations lack sufficient spatial and semantic refinement to provide meaningful order supervision.

\subsection{Physical Spatial Aware Data Augmentation}
\label{sec:data_generation}

In real-world scenarios, document images captured by mobile devices exhibit complex geometric distortions that can be categorized into two distinct physical processes: intrinsic surface deformation and extrinsic projective transformation. 
However, standard document datasets typically provide only axis-aligned bounding boxes or flat polygon annotations, which fail to capture the pixel-level nuances of warped pages. 

\begin{figure}[ht]
\centering
\includegraphics[width=\linewidth]{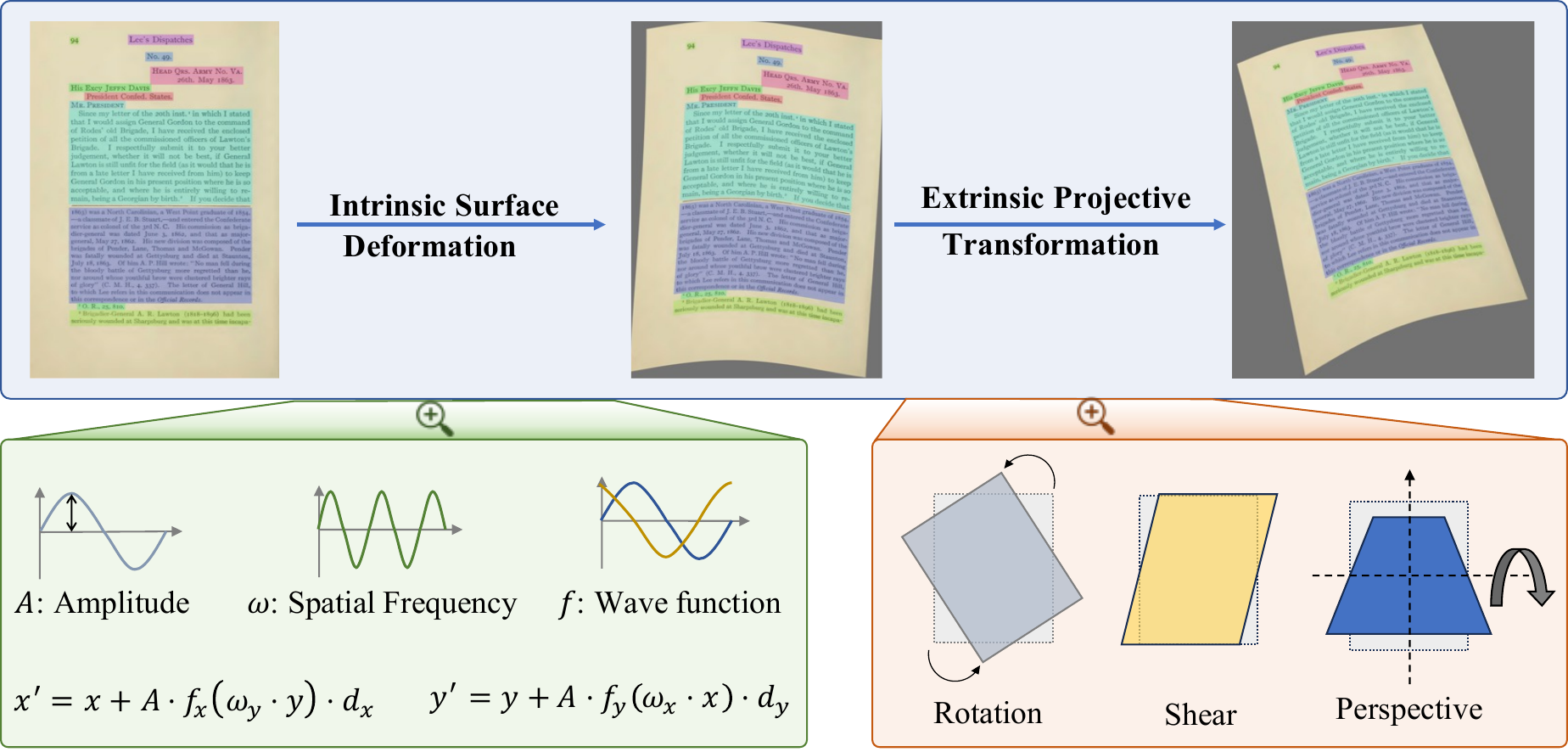}
\caption{The Physical Spatial Aware Data Augmentation Pipeline.}
\label{fig:DataAug}
\end{figure}

To bridge this domain gap, we propose an online physical spatial aware data augmentation pipeline, as shown on Fig.~\ref{fig:DataAug}, that decouples the augmentation into a cascaded process: first simulating the non-rigid physical warping of the paper surface, and subsequently modeling the stochastic camera viewpoints. 
This pipeline dynamically synthesizes realistic distortions from standard flat annotations, providing a robust foundation for training deformation-aware segmentation models. 

\textbf{Stage 1: Intrinsic Surface Deformation (ISD).}
To characterize the non-rigid geometric priors of physical documents, such as page curling in bound volumes. We propose a mesh-based warping operator. Unlike standard linear transformations, ISD employs smooth, periodic displacement fields to simulate paper deformation. For each pixel $(x, y)$, the deformed coordinates $(x', y')$ are computed as:

\begin{equation}
    x' = x + A \cdot f_x\!\left(\omega_y \cdot y\right) \cdot d_x, \qquad
    y' = y + A \cdot f_y\!\left(\omega_x \cdot x\right) \cdot d_y
\end{equation}

where $f_x, f_y \in \{\sin, \cos\}$ are randomly sampled wave functions, $A = \alpha \min(H, W)$ is the displacement amplitude proportional to the image resolution with $\alpha$ drawn from a configurable range, $\omega_x = \beta\pi / W$ and $\omega_y = \beta\pi / H$ control the spatial frequency with $\beta$ sampled uniformly, and $d_x, d_y \in \{-1, 1\}$ randomize the displacement direction. The operator supports two stochastic modes: a single-axis mode that applies displacement along only one axis to simulate unidirectional page curl, and an intense mode that applies displacement along both axes simultaneously to simulate compound deformations. Image and masks are remapped with bilinear and nearest-neighbor interpolation, respectively. Border padding is dynamically computed from the maximum displacement amplitudes to preserve all content.

\textbf{Stage 2: Extrinsic Projective Transformation (EPT).}
Document images frequently undergo geometric distortions due to variations in camera perspective. To account for this, EPT simulates the variability of camera viewpoints in uncontrolled environments. We formulate this process as a composite homography $M$, which maps the previously warped surface into a 3D-aware perspective space. The transformation matrix $M \in \mathbb{R}^{3 \times 3}$ is decomposed as follows:

\begin{equation}
    M = S \cdot R \cdot P
\end{equation}

where $P$ introduces perspective foreshortening via its projective parameters $p_{20}$ and $p_{21}$, mimicking oblique viewing angles; $R$ and $S$ account for in-plane rotation and angular shear, respectively, to simulate natural hand-held capture. Finally, the instance-level ground truth is updated by re-extracting axis-aligned bounding boxes from the transformed pixel-accurate masks.

\section{Experiments}
\label{sec:experiments}

\subsection{Experimental Setup}
\label{subsec:experimental_setup}

\subsubsection{Datasets}
\label{subsubsec:dataset}

To ensure robust model performance across diverse real-world document scenarios, we curate an in-house dataset for layout analysis. The data sources encompass 38k document images across diverse domains, including Academic papers, Textbooks, Market Analysis, Financial reports, Slides, Newspapers, Supplementary Teaching Materials, Examination Papers, and various Invoices and Receipts. The dataset features meticulous manual annotations across 25 distinct layout element categories: Paragraph Title, Image, Text, Number, Abstract, Content, Figure Title, Display Formula, Table, Reference, Doc Title, Footnote, Header, Algorithm, Footer, Seal, Chart, Formula Number, Aside Text, Reference Content, Header Image, Footer Image, Inline Formula, Vertical Text, and Vision Footnote. All documents are manually annotated with element-level boundaries and their corresponding reading order, enabling effective training and evaluation for both layout element detection and reading order prediction. This high-quality ground truth ensures that the model can accurately reconstruct both the spatial structure and the logical flow of complex documents. All experiments in this paper are trained on this dataset. For evaluation, we adopt OmniDocBench v1.5~\cite{ouyang2025omnidocbenchbenchmarkingdiversepdf}and its real-world document–oriented derivative dataset, Real5-OmniDocBench~\cite{zhou2026real5omnidocbenchfullscalephysicalreconstruction}.

\subsubsection{Evaluation Metrics}
\label{subsubsec:evaluation_metrics}

Due to the lack of standardized evaluation metrics that comprehensively measure layout analysis capability—covering both layout detection and reading order prediction—we evaluate RT-DocLayout using downstream document parsing benchmarks. Rather than relying solely on geometric detection accuracy, this protocol assesses whether predicted layouts effectively support structured document parsing in realistic scenarios.

Specifically, we conduct evaluation on OmniDocBench v1.5 and its real-world extension, Real5-OmniDocBench, measuring performance on formulas, tables, textual content, and reading order. These metrics jointly reflect spatial localization accuracy as well as the structural and semantic coherence captured by the layout model.

\subsubsection{Implementation Details}
\label{subsubsec:implementation_details}

The model is initialized with the pre-trained weights of PP-DocLayout\_plus-L, we scaled our training corpus to over 38k high-quality document samples. Each sample underwent rigorous manual annotation to provide ground truth, including coordinates, categorical labels, and absolute reading order for every layout elements. To achieve the environmental robustness, we designed a specialized Distortion-Aware Data Augmentation pipeline. Unlike standard augmentations, this pipeline specifically simulates complex physical deformations found in real-world mobile photography. We utilize the AdamW optimizer with a weight decay of 0.0001. The learning rate is set to a constant \(2 \times 10^{-4}\) to ensure stable convergence of the integrated antisymmetric pairwise scorer and Mask heads. The model is trained for 150 epochs with a total batch size of 32.

\begin{table*}[t]
\centering
\caption{Comparison of different layout analysis modules combined with various recognition systems on OmniDocBench v1.5. As DocLayout-YOLO does not output reading order, we employ the XYCut algorithm as a post-processing.}
\renewcommand{\arraystretch}{1.2}
\setlength{\tabcolsep}{3pt}
\resizebox{\textwidth}{!}{
\begin{tabular}{l|c|c|c|c|c c c c c}
\toprule
\multicolumn{1}{c|}{\textbf{Layout Analysis}}
& \multirow{1}{*}{\textbf{Params}}
& \multirow{1}{*}{\textbf{FPS}}
& \multirow{1}{*}{\textbf{Recognition Method}}
& \multirow{1}{*}{\textbf{Overall}$\uparrow$}
& \multirow{1}{*}{\textbf{Text\textsuperscript{Edit}}$\downarrow$}
& \multirow{1}{*}{\textbf{Formula\textsuperscript{CDM}}$\uparrow$}
& \multirow{1}{*}{\textbf{Table\textsuperscript{TEDS}}$\uparrow$}
& \multirow{1}{*}{\textbf{Reading Order\textsuperscript{Edit}}$\downarrow$} \\
\midrule
\multirow{4}{*}{DocLayout-YOLO~\cite{zhao2024doclayoutyoloenhancingdocumentlayout}}
& \multirow{4}{*}{20M}
& \multirow{4}{*}{107.4}
& MonkeyOCR~\cite{li2026monkeyocrdocumentparsingstructurerecognitionrelation} & 85.10 & 0.130 & 85.99 & 82.31 & 0.169 \\
& & & MinerU2.5~\cite{niu2025mineru25decoupledvisionlanguagemodel} & 85.73 & 0.138 & 87.26 & 83.73 & 0.177 \\
& & & Dolphin-v2~\cite{feng2026dolphinv2universaldocumentparsing} & 84.14 & 0.150 & 83.92 & 83.47 & 0.178 \\
& & & PaddleOCR-VL-1.5-0.9B~\cite{cui2026paddleocrvl15multitask09bvlm} & 87.29 & 0.127 & 88.74 & 85.83 & 0.169 \\
\midrule
\multirow{4}{*}{MinerU2.5~\cite{niu2025mineru25decoupledvisionlanguagemodel}}
& \multirow{4}{*}{1.2B}
& \multirow{4}{*}{2.4}
& MonkeyOCR~\cite{li2026monkeyocrdocumentparsingstructurerecognitionrelation} & 89.69 & 0.047 & 87.06 & 86.71 & 0.050 \\
& & & MinerU2.5~\cite{niu2025mineru25decoupledvisionlanguagemodel} & 90.17 & 0.052 & 87.81 & 87.95 & 0.053 \\
& & & Dolphin-v2~\cite{feng2026dolphinv2universaldocumentparsing} & 88.25 & 0.068 & 83.55 & 88.02 & 0.062 \\
& & & PaddleOCR-VL-1.5-0.9B~\cite{cui2026paddleocrvl15multitask09bvlm} & 91.31 & 0.046 & 87.57 & 90.92 & 0.051 \\
\midrule
\multirow{4}{*}{Dolphin-v2~\cite{feng2026dolphinv2universaldocumentparsing}}
& \multirow{4}{*}{3B}
& \multirow{4}{*}{0.9}
& MonkeyOCR~\cite{li2026monkeyocrdocumentparsingstructurerecognitionrelation} & 89.14 & 0.051 & 87.28 & 85.24 & 0.061 \\
& & & MinerU2.5~\cite{niu2025mineru25decoupledvisionlanguagemodel} & 89.58 & 0.055 & 87.71 & 86.57 & 0.067 \\
& & & Dolphin-v2~\cite{feng2026dolphinv2universaldocumentparsing} & 87.49 & 0.073 & 83.04 & 86.69 & 0.067 \\
& & & PaddleOCR-VL-1.5-0.9B~\cite{cui2026paddleocrvl15multitask09bvlm} & 90.36 & 0.049 & 87.60 & 88.36 & 0.060 \\
\midrule
\multirow{4}{*}{PP-DocLayoutV2~\cite{cui2025paddleocrvlboostingmultilingualdocument}}
& \multirow{4}{*}{53M}
& \multirow{4}{*}{110.9}
& MonkeyOCR~\cite{li2026monkeyocrdocumentparsingstructurerecognitionrelation} & 91.32 & 0.038 & 89.41 & 88.41 & 0.045 \\ \
& & & MinerU2.5~\cite{niu2025mineru25decoupledvisionlanguagemodel} & 92.00 & 0.042 & 90.51 & 89.66 & 0.051 \\
& & & Dolphin-v2~\cite{feng2026dolphinv2universaldocumentparsing} & 87.49 & 0.073 & 83.04 & 86.69 & 0.067 \\
& & & PaddleOCR-VL-1.5-0.9B~\cite{cui2026paddleocrvl15multitask09bvlm} & 93.56 & \textbf{0.034} & 92.09 & 92.03 & 0.042 \\
\midrule
 \multirow{4}{*}{\textbf{RT-DocLayout}}
& \multirow{4}{*}{33M}
& \multirow{4}{*}{132.1}
& MonkeyOCR~\cite{li2026monkeyocrdocumentparsingstructurerecognitionrelation} & 92.51 & 0.039 & 92.37 & 89.07 & 0.044 \\ 
& & & MinerU2.5~\cite{niu2025mineru25decoupledvisionlanguagemodel} & 92.68 & 0.043 & 92.19 & 90.12 & 0.049 \\
& & & Dolphin-v2~\cite{feng2026dolphinv2universaldocumentparsing} & 90.72 & 0.060 & 89.17 & 89.00 & 0.058 \\
& & & PaddleOCR-VL-1.5-0.9B~\cite{cui2026paddleocrvl15multitask09bvlm} & \textbf{94.50} & 0.035 & \textbf{94.21} & \textbf{92.76} & \textbf{0.042} \\
\bottomrule
\end{tabular}
}
\label{tab:layout_recognition_combo}
\end{table*}

\subsection{Main Results}
\label{subsec:main_results}

\subsubsection{OmniDocBench v1.5}

Table~\ref{tab:layout_recognition_combo} presents the performance of different layout analysis modules combined with various recognition methods on OmniDocBench v1.5.

Across all evaluated recognition methods, RT-DocLayout achieves the best Overall performance. When paired with PaddleOCR-VL-1.5-0.9B, RT-DocLayout reaches an Overall score of 94.50, which is the highest among all evaluated combinations. Consistent improvements can also be observed when RT-DocLayout is combined with MonkeyOCR, MinerU2.5, and Dolphin-v2, demonstrating its compatibility with heterogeneous recognition architectures.

In addition to the Overall metric, RT-DocLayout brings consistent gains across multiple evaluation dimensions, including text accuracy (Edit distance), formula recognition (CDM), table structure similarity (TEDS), and reading order prediction (Edit distance). These results indicate that improving layout analysis quality can positively influence recognition performance in a holistic manner.

Moreover, RT-DocLayout contains only 33 M parameters and infer at 132.1 FPS, achieving both the highest inference speed and the best overall accuracy among all evaluated layout analysis modules.

\begin{table*}[t]
\centering
\caption{Comparison of layout analysis and recognition system combinations on Real5-OmniDocBench. As DocLayout-YOLO does not output reading order, we employ the XYCut algorithm as a post-processing. Overall is the average across five scenarios---Scanning, Warping, Screen-Photography, Illumination, and Skew (abbreviated as Scan, Warp, Screen, Illum, Skew).}
\renewcommand{\arraystretch}{1.2}
\setlength{\tabcolsep}{3pt}
\resizebox{\textwidth}{!}{
\begin{tabular}{l|c|c|c|c|c c c c c}
\toprule
\multicolumn{1}{c|}{\textbf{Layout Analysis}}
& \multirow{1}{*}{\textbf{Params}}
& \multirow{1}{*}{\textbf{FPS}}
& \multirow{1}{*}{\textbf{Recognition Method}}
& \multirow{1}{*}{\textbf{Overall}$\uparrow$}
& \multirow{1}{*}{\textbf{Scan}$\uparrow$}
& \multirow{1}{*}{\textbf{Warp}$\uparrow$}
& \multirow{1}{*}{\textbf{Screen}$\uparrow$}
& \multirow{1}{*}{\textbf{Illum}$\uparrow$}
& \multirow{1}{*}{\textbf{Skew}$\uparrow$}
 \\
\midrule
\multirow{4}{*}{DocLayout-YOLO~\cite{zhao2024doclayoutyoloenhancingdocumentlayout}}
& \multirow{4}{*}{20M}
& \multirow{4}{*}{107.4}
& MonkeyOCR~\cite{li2026monkeyocrdocumentparsingstructurerecognitionrelation} & 73.57 & 84.09 & 74.81 & 76.89 & 81.99 & 50.06 \\
& & & MinerU2.5~\cite{niu2025mineru25decoupledvisionlanguagemodel} & 72.92 & 85.00 & 72.92 & 77.38 & 82.68 & 46.62 \\
& & & Dolphin-v2~\cite{feng2026dolphinv2universaldocumentparsing} & 72.72 & 83.40 & 73.59 & 75.90 & 81.00 & 49.71 \\
& & & PaddleOCR-VL-1.5-0.9B~\cite{cui2026paddleocrvl15multitask09bvlm} & 74.07 & 85.57 & 74.70 & 77.67 & 83.65 & 48.74 \\
\midrule
\multirow{4}{*}{MinerU2.5~\cite{niu2025mineru25decoupledvisionlanguagemodel}}
& \multirow{4}{*}{1.2B}
& \multirow{4}{*}{2.4}
& MonkeyOCR~\cite{li2026monkeyocrdocumentparsingstructurerecognitionrelation} & 85.23 & 88.70 & 84.10 & 87.16 & 87.63 & 78.56 \\
& & & MinerU2.5~\cite{niu2025mineru25decoupledvisionlanguagemodel} & 84.55 & 88.37 & 82.85 & 88.29 & 88.02 & 75.20 \\
& & & Dolphin-v2~\cite{feng2026dolphinv2universaldocumentparsing} & 82.48 & 87.20 & 82.72 & 85.89 & 86.17 & 70.45 \\
& & & PaddleOCR-VL-1.5-0.9B~\cite{cui2026paddleocrvl15multitask09bvlm} & 85.67 & 89.74 & 83.92 & 89.24 & 89.29 & 76.16 \\
\midrule
\multirow{4}{*}{Dolphin-v2~\cite{feng2026dolphinv2universaldocumentparsing}}
& \multirow{4}{*}{3B}
& \multirow{4}{*}{0.9}
& MonkeyOCR~\cite{li2026monkeyocrdocumentparsingstructurerecognitionrelation} & 69.74 & 87.86 & 58.41 & 81.01 & 68.31 & 53.12 \\
& & & MinerU2.5~\cite{niu2025mineru25decoupledvisionlanguagemodel} & 60.88 & 87.70 & 45.38 & 76.75 & 58.54 & 36.01 \\
& & & Dolphin-v2~\cite{feng2026dolphinv2universaldocumentparsing} & 67.40 & 85.72 & 58.08 & 73.30 & 64.03 & 55.84 \\
& & & PaddleOCR-VL-1.5-0.9B~\cite{cui2026paddleocrvl15multitask09bvlm} & 67.66 & 88.66 & 52.58 & 81.06 & 66.15 & 49.84 \\
\midrule
\multirow{4}{*}{PP-DocLayoutV2~\cite{cui2025paddleocrvlboostingmultilingualdocument}}
& \multirow{4}{*}{53M}
& \multirow{4}{*}{110.9}
& MonkeyOCR~\cite{li2026monkeyocrdocumentparsingstructurerecognitionrelation} & 85.10 & 90.47 & 85.90 & 81.60 & 88.86 & 78.65 \\
& & & MinerU2.5~\cite{niu2025mineru25decoupledvisionlanguagemodel} & 83.87 & 90.98 & 83.88 & 81.98 & 89.06 & 73.44 \\
& & & Dolphin-v2~\cite{feng2026dolphinv2universaldocumentparsing} & 84.16 & 89.46 & 84.54 & 81.11 & 87.97 & 77.71 \\
& & & PaddleOCR-VL-1.5-0.9B~\cite{cui2026paddleocrvl15multitask09bvlm} & 85.83 & 92.04 & 86.00 & 83.28 & 90.80 & 77.03 \\
\midrule
\multirow{4}{*}{\textbf{RT-DocLayout}}
& \multirow{4}{*}{33M}
& \multirow{4}{*}{132.1}
& MonkeyOCR~\cite{li2026monkeyocrdocumentparsingstructurerecognitionrelation} & 89.79 & 91.73 & 89.38 & 89.67 & 89.72 & 88.44 \\
& & & MinerU2.5~\cite{niu2025mineru25decoupledvisionlanguagemodel} & 89.89 & 91.74 & 88.96 & 90.38 & 90.31 & 88.06 \\
& & & Dolphin-v2~\cite{feng2026dolphinv2universaldocumentparsing} & 88.97 & 90.36 & 88.47 & 88.61 & 89.39 & 88.04 \\
& & & PaddleOCR-VL-1.5-0.9B~\cite{cui2026paddleocrvl15multitask09bvlm} &\textbf{ 92.05} & \textbf{93.43} &\textbf{ 91.25} & \textbf{91.76} & \textbf{92.16} & \textbf{91.66} \\
\bottomrule
\end{tabular}
}
\label{tab:layout_recognition_combo_v2}
\end{table*}

\subsubsection{Real5-OmniDocBench}
To further evaluate robustness under real-world physical distortions, we conduct experiments on Real5-OmniDocBench, which includes five challenging scenarios: Scanning, Warping, Screen-Photography, Illumination, and Skew.

As shown in Table~\ref{tab:layout_recognition_combo_v2}, RT-DocLayout consistently achieves superior performance across different recognition systems. When combined with PaddleOCR-VL-1.5-0.9B, it obtains an Overall score of 92.05\%, which is the highest among all configurations.

The improvements are particularly evident in geometrically challenging scenarios such as Warping and Skew. For example, when paired with PaddleOCR-VL-1.5-0.9B, RT-DocLayout achieves 91.25\% on Warp and 91.66\% on Skew. In contrast, box-based layout analysis methods exhibit more noticeable performance degradation in these settings. This observation suggests that more precise geometric modeling can improve robustness under various distortions.

Furthermore, the performance gains brought by RT-DocLayout remain consistent across different recognition methods, indicating that the improvements are not specific to a particular downstream model but stem from enhanced layout analysis.

\subsection{Ablation Study and Analysis}
\label{subsec:ablation_study_and_analysis}

\begin{table}[t]
\centering
\caption{
    Ablation study of our model components on OmniDocBench v1.5 and its real-world degradation subset, Real5-OmniDocBench. 
    We analyze the impact of the localization method (Loc: bbox vs. mask), the reading order module (Order), and our proposed data augmentation (DA). 
    Metrics are Overall accuracy (O, $\uparrow$) and Reading Order distance (RO, $\downarrow$). 
    Our full model (last row) achieves the best performance, demonstrating the effectiveness of each component.
}
\renewcommand{\arraystretch}{1.15} 
\setlength{\tabcolsep}{3pt}      

\resizebox{\columnwidth}{!}{
\begin{tabular}{ccc|cc|*{5}{c@{\hskip 0.5em}c}}
\toprule
\multicolumn{3}{c|}{\multirow{2}{*}{\textbf{Components}}} &
\multicolumn{2}{c|}{\multirow{2}{*}{\textbf{OmniDocBench v1.5}}} &
\multicolumn{10}{c}{\textbf{Real5-OmniDocBench}} \\
\cmidrule(lr){6-15}
& & & & &
\multicolumn{2}{c}{\textbf{Scan}} &
\multicolumn{2}{c}{\textbf{Warp}} &
\multicolumn{2}{c}{\textbf{Screen}} &
\multicolumn{2}{c}{\textbf{Illum}} &
\multicolumn{2}{c}{\textbf{Skew}} \\
\cmidrule(lr){1-3} \cmidrule(lr){4-5} \cmidrule(lr){6-7} \cmidrule(lr){8-9} \cmidrule(lr){10-11} \cmidrule(lr){12-13} \cmidrule(lr){14-15}
\textbf{Loc} & \textbf{Order} & \textbf{DA} &
\textbf{O}$\uparrow$ & \textbf{RO}$\downarrow$ &
\textbf{O}$\uparrow$ & \textbf{RO}$\downarrow$ &
\textbf{O}$\uparrow$ & \textbf{RO}$\downarrow$ &
\textbf{O}$\uparrow$ & \textbf{RO}$\downarrow$ &
\textbf{O}$\uparrow$ & \textbf{RO}$\downarrow$ &
\textbf{O}$\uparrow$ & \textbf{RO}$\downarrow$ \\
\midrule
bbox & XYCut~\cite{ha1995recursive} & $\times$
  & 93.81 & 0.082
  & 90.30 & 0.749 & 84.44 & 0.720 & 88.35 & 0.744 & 88.90 & 0.747 & 77.18 & 0.662 \\
bbox+mask & XYCut~\cite{ha1995recursive} & $\times$
  & 94.33 & 0.074
  & 90.22 & 0.742 & 84.40 & 0.721 & 88.55 & 0.744 & 89.14 & 0.747 & 75.63 & 0.673 \\
bbox+mask & Decoupled & $\times$
  & 93.87 & 0.189
  & 92.85 & 0.192 & 85.91 & 0.238 & 91.22 & 0.195 & 91.55 & 0.188 & 76.24 & 0.325 \\
bbox+mask & Coupled & $\times$
  & 94.39 & \textbf{0.041}
  & 93.29 & 0.045 & 86.68 & 0.092 & 91.55 & \textbf{0.050} & 92.12 & 0.054 & 77.55 & 0.146 \\
\midrule
\textbf{bbox+mask} & \textbf{Coupled} & \textbf{\checkmark}
  & \textbf{94.50} &0.042
  & \textbf{93.43} & \textbf{0.045} & \textbf{91.25} & \textbf{0.063} & \textbf{91.76} & 0.059 & \textbf{92.16} & \textbf{0.051} & \textbf{91.66} & \textbf{0.061} \\
\bottomrule
\end{tabular}
}
\label{tab:layout_ablation_study}
\end{table}

We conduct a structured ablation study on OmniDocBench v1.5 and Real5-OmniDocBench, to analyze how each component of RT-DocLayout contributes to unified document layout analysis. All experiments use PaddleOCR-VL-1.5-0.9B as the downstream recognizer. The results are summarized in Table~\ref{tab:layout_ablation_study}.

\textbf{Effectiveness of Mask-based Localization.}
As shown in the first two rows of Table~\ref{tab:layout_ablation_study}, replacing pure bounding-box supervision with joint mask prediction ($bbox+mask$) improves the Overall accuracy ($O$) from 93.81\% to 94.33\% on OmniDocBench v1.5. The introduction of pixel-level masks provides finer structural cues, allowing the model to better distinguish overlapping or closely-packed elements, which is reflected in the consistent $O$ gains across most scenarios on Real5-OmniDocBench.


\textbf{Coupled Reading Order Learning.}
A key contribution of RT-DocLayout is the end-to-end prediction of reading order. Compared to the heuristic XYCut~\cite{ha1995recursive}, our Coupled learning approach (penultimate row) significantly reduces the Reading Order distance ($RO$) from 0.074 to 0.041 on OmniDocBench v1.5. More importantly, in complex scenarios like the "Warp" subset, the $RO$ distance drops drastically from 0.721 (XYCut) to 0.092. It is worth noting that the Reading Order distance ($RO$) of decoupled learning strategy (reading order as a separate trained model without joint optimization) is 0.189, performing significantly worse than 0.074 (XYCut) on OmniDocBench v1.5. However, this performance deficit is effectively reversed through joint optimization. This further validates that integrating reading order directly into the layout perception phase allows the model to capture global structural dependencies more effectively.

\textbf{Robustness through realistic augmentation.}
The most significant performance leap on Real5-OmniDocBench comes from our proposed data augmentation (DA). When DA is enabled (last row), the model's performance on the "Warp" and "Skew" subsets of Real5-OmniDocBench increases by 4.57 (86.68\% $\to$ 91.25\%) and 14.11 percentage points (77.55\% $\to$ 91.66\%) respectively. Crucially, this robustness extends beyond localization to reading order: on the "Skew" subset, the Reading Order (RO) distance drops significantly from 0.146 to 0.061, while on the "Warp" subset, it improves from 0.092 to 0.063. Since our model generates mask-level outputs, the DA can simulate realistic geometric distortions (like page curling or camera tilting) while maintaining pixel-level ground truth. 
These results demonstrate that by training on mask-guided synthetic distortions, RT-DocLayout learns to preserve spatial-logical consistency even under non-linear transformations. Consequently, our model achieves state-of-the-art robustness against real-world physical deformations without sacrificing accuracy on OmniDocBench v1.5.


\begin{figure*}[!htpb]
  \centering
  \subfloat[\label{fig:qual_a}]{%
    \begin{minipage}{0.48\linewidth}
      \centering
      \includegraphics[width=\linewidth]{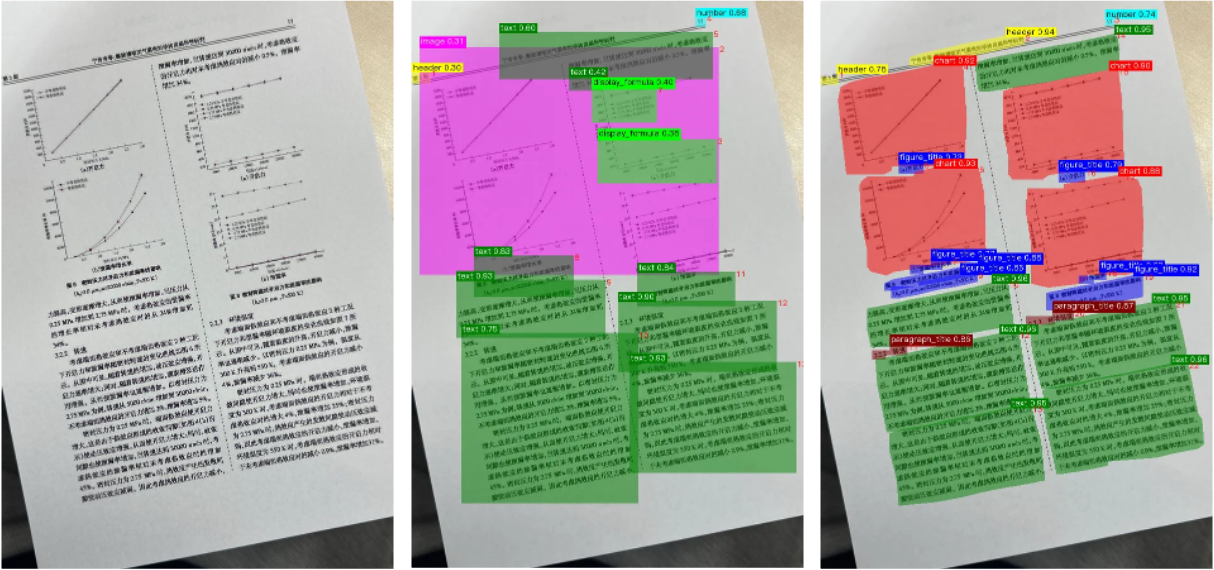}\\[2pt]
      \makebox[0.33\linewidth][c]{\scriptsize Origin}%
      \makebox[0.33\linewidth][c]{\scriptsize PP-DocLayoutV2}%
      \makebox[0.33\linewidth][c]{\scriptsize\textbf{Ours}}
    \end{minipage}%
  }%
  \hfill
  \subfloat[\label{fig:qual_b}]{%
    \begin{minipage}{0.48\linewidth}
      \centering
      \includegraphics[width=\linewidth]{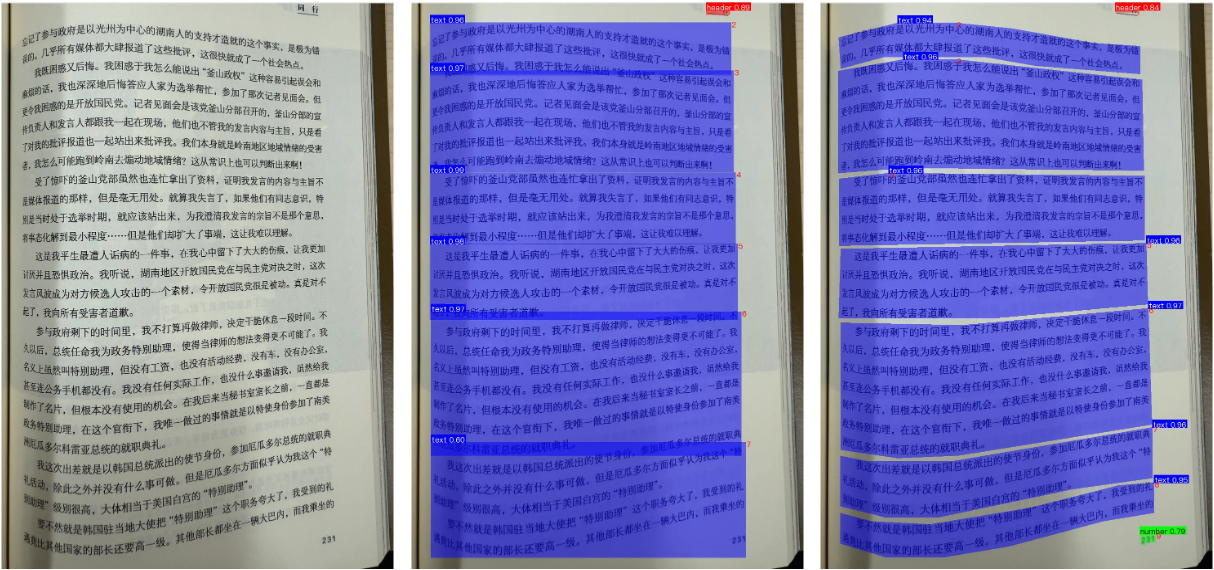}\\[2pt]
      \makebox[0.33\linewidth][c]{\scriptsize Origin}%
      \makebox[0.33\linewidth][c]{\scriptsize PP-DocLayoutV2}%
      \makebox[0.33\linewidth][c]{\scriptsize\textbf{Ours}}
    \end{minipage}%
  }%
  \\[4pt]
  \subfloat[\label{fig:qual_c}]{%
    \begin{minipage}{0.48\linewidth}
      \centering
      \includegraphics[width=\linewidth]{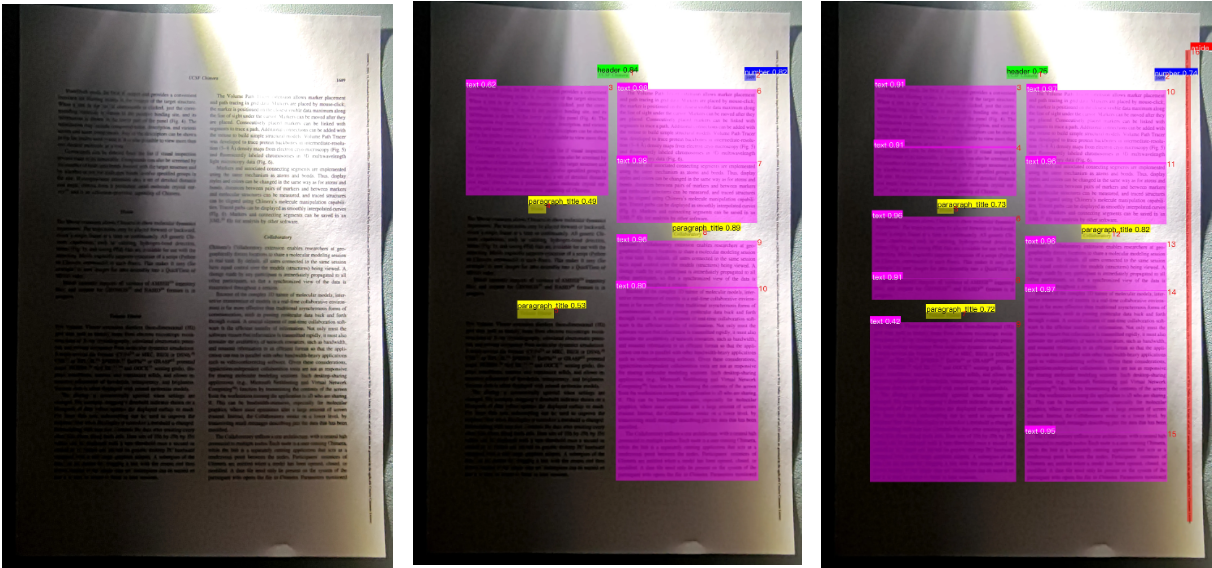}\\[2pt]
      \makebox[0.33\linewidth][c]{\scriptsize Origin}%
      \makebox[0.33\linewidth][c]{\scriptsize PP-DocLayoutV2}%
      \makebox[0.33\linewidth][c]{\scriptsize\textbf{Ours}}
    \end{minipage}%
  }%
  \hfill
  \subfloat[\label{fig:qual_d}]{%
    \begin{minipage}{0.48\linewidth}
      \centering
      \includegraphics[width=\linewidth]{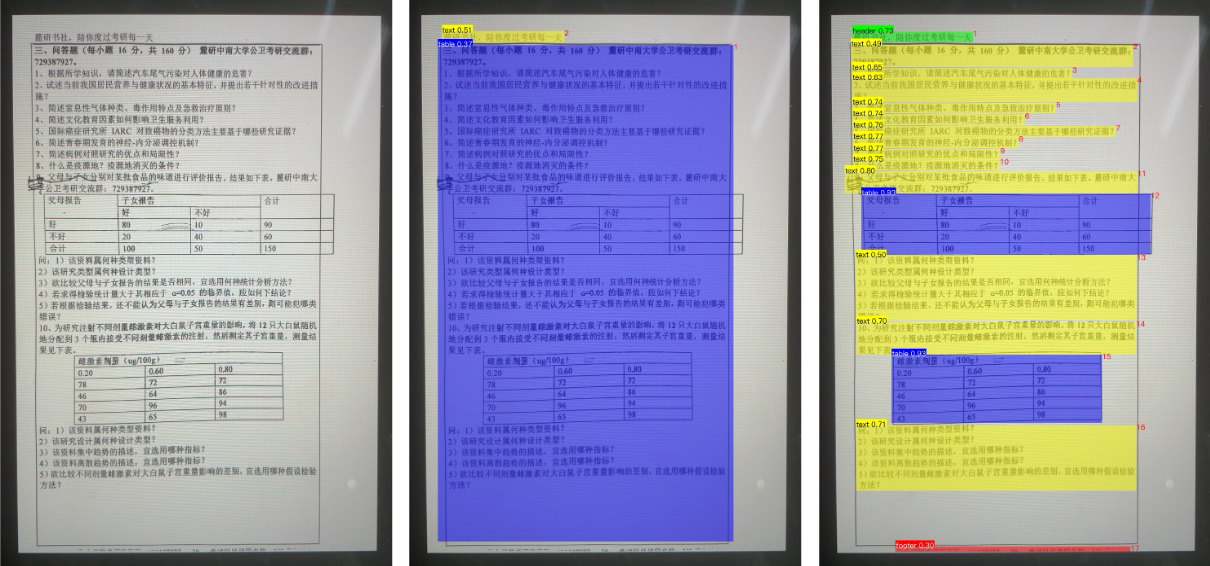}\\[2pt]
      \makebox[0.33\linewidth][c]{\scriptsize Origin}%
      \makebox[0.33\linewidth][c]{\scriptsize PP-DocLayoutV2}%
      \makebox[0.33\linewidth][c]{\scriptsize\textbf{Ours}}
    \end{minipage}%
  }%

  \caption{
    Qualitative comparison on four real-world scenarios. Each group shows, left to right: Origin, PP-DocLayoutV2, \textbf{Ours}.
    \textbf{(a)}~Skew. \textbf{(b)}~Page warping. \textbf{(c)}~Non-uniform illumination. \textbf{(d)}~Screen photography.
  }
  \label{fig:qualitative}
\end{figure*}

\subsection{Qualitative Analysis and Case Insights}
\label{subsec:qualitative_analysis_and_case_insights}

Fig.~\ref{fig:qualitative} compares RT-DocLayout with PP-DocLayoutV2, a previous state-of-the-art real-time document layout analysis method, across four real-world scenarios: skew~(a), page warping~(b), non-uniform illumination~(c), and screen photography~(d). In (a) and (b), axis-aligned boxes span background and neighboring elements when the document geometry deviates from a fronto-parallel plane; RT-DocLayout masks follow the actual boundaries under each distortion. In (c), PP-DocLayoutV2 misses content in the darker region due to low contrast; RT-DocLayout produces consistent predictions across the full layout. In (d), PP-DocLayoutV2 conflates tables, text blocks, and titles into coarse regions; RT-DocLayout recovers fine-grained element boundaries under the screen-captured condition.These cases demonstrate that RT-DocLayout produces accurate and geometrically consistent layout predictions across diverse real-world degradations.

\section{Conclusion}
\label{sec:conclusion}

In this paper, we presented RT-DocLayout, a unified and efficient Transformer-based framework for real-time document layout analysis. Moving beyond the conventional box-centric paradigm, RT-DocLayout unifies layout detection, pixel-level segmentation, and reading order prediction within a single non-autoregressive forward pass. This unified formulation improves structural consistency and geometric robustness, particularly for documents with complex distortions.

Extensive experiments show that RT-DocLayout, with only 33 M parameters, achieves a superior balance between accuracy and efficiency, outperforming substantially larger models in both geometric fidelity and real-time deployment capability. RT-DocLayout establishes a scalable and structurally aware foundation for next-generation document intelligence systems.

\bibliography{main}

\setcounter{figure}{0}
\makeatletter 
\renewcommand{\thefigure}{A\@arabic\c@figure}
\makeatother

\setcounter{table}{0}
\makeatletter 
\renewcommand{\thetable}{A\@arabic\c@table}
\makeatother

\clearpage 
\newpage

\end{document}